\documentclass[letterpaper, 10 pt, conference]{ieeeconf}  

\IEEEoverridecommandlockouts                              

\usepackage{graphics} 
\usepackage{epsfig} 
\usepackage{mathptmx} 
\usepackage{times} 
\usepackage{amsmath} 
\usepackage{amssymb}  
\usepackage{graphicx}
\usepackage{subcaption}
\usepackage{booktabs}       

\usepackage{floatrow}
\newfloatcommand{capbtabbox}{table}[][\FBwidth]
\usepackage{caption}
\usepackage[utf8]{inputenc} 
\usepackage[T1]{fontenc}    
\usepackage{hyperref}       
\usepackage{url}            
\usepackage{booktabs}       
\usepackage{amsfonts}       
\usepackage{nicefrac}       
\usepackage{microtype}      
\usepackage{amsmath}
\usepackage{graphicx}
\usepackage{microtype}
\usepackage{subcaption}
\usepackage{algorithm}
\usepackage{algorithmic}

\title{\LARGE \bf Hypothesis-Driven Skill Discovery for\\ Hierarchical Deep Reinforcement Learning}

\author{Caleb Chuck$^{1}$, Supawit Chockchowwat$^{1}$ and Scott Niekum$^{1}$
\thanks{$^{1}$The University of Texas at Austin Personal Robotics and Automation Lab. Contact: {\tt\small calebc@cs.utexas.edu}}%
}

\begin{document}

\maketitle
\thispagestyle{empty}
\pagestyle{empty}
\vskip -2cm

\begin{abstract}

Deep reinforcement learning (DRL) is capable of learning high-performing policies on a variety of complex high-dimensional tasks, ranging from video games to robotic manipulation. However, standard DRL methods often suffer from poor sample efficiency, partially because they aim to be entirely problem-agnostic. In this work, we introduce a novel approach to exploration and hierarchical skill learning that derives its sample efficiency from intuitive assumptions it makes about the behavior of objects both in the physical world and simulations which mimic physics. Specifically, we propose the Hypothesis Proposal and Evaluation (HyPE) algorithm, which discovers objects from raw pixel data, generates hypotheses about the controllability of observed changes in object state, and learns a hierarchy of skills to test these hypotheses.
We demonstrate that HyPE can dramatically improve the sample efficiency of policy learning in two different domains: a simulated robotic block-pushing domain, and a popular benchmark task: Breakout. In these domains, HyPE learns high-scoring policies an order of magnitude faster than several state-of-the-art reinforcement learning methods.

\end{abstract}


\section{Introduction}

While recent advances in deep reinforcement learning (DRL) have been used to obtain exciting results on a variety of high-dimensional visual tasks, these algorithms often require large amounts of data in order to achieve good performance. In real-world domains such as robotics, this data is difficult and expensive to collect in sufficient quantity. When using high dimensional observations like images as input state, a natural factorization of state often exists that reduces the state space complexity. This factorization reduces the space of pixels to a space where only sparse instances of interaction occur \cite{review}---for example, a key unlocking a door, a bat hitting a baseball, or gripper-object contact in robotic manipulation. Such interactions often create state-space bottlenecks \cite{csimcsek2009skill}---a small subset of states which must be reached in order for the agent to access large regions of the state space. This feature of the state space can make exploration difficult, but can also present opportunities for hierarchical RL algorithms \cite{bacon, konidaris, vezhnevets, sharma2020dynamics} to learn \textit{options} \cite{options} that efficiently navigate between regions separated by bottlenecks.
Unfortunately, hierarchical RL also often learns slowly because extrinsic reward must be experienced, often repeatedly, before the agent can start to learn meaningful behavior or skills. For sufficiently difficult sparse-reward tasks, even with state-of-the-art exploration methods \cite{ostrovski2017count, pathak2017curiosity, hester}, the agent may take an exceptionally long time to see a positive reward even once.

\begin{figure}
\begin{center}
\centerline{\includegraphics[width=1.0\columnwidth]{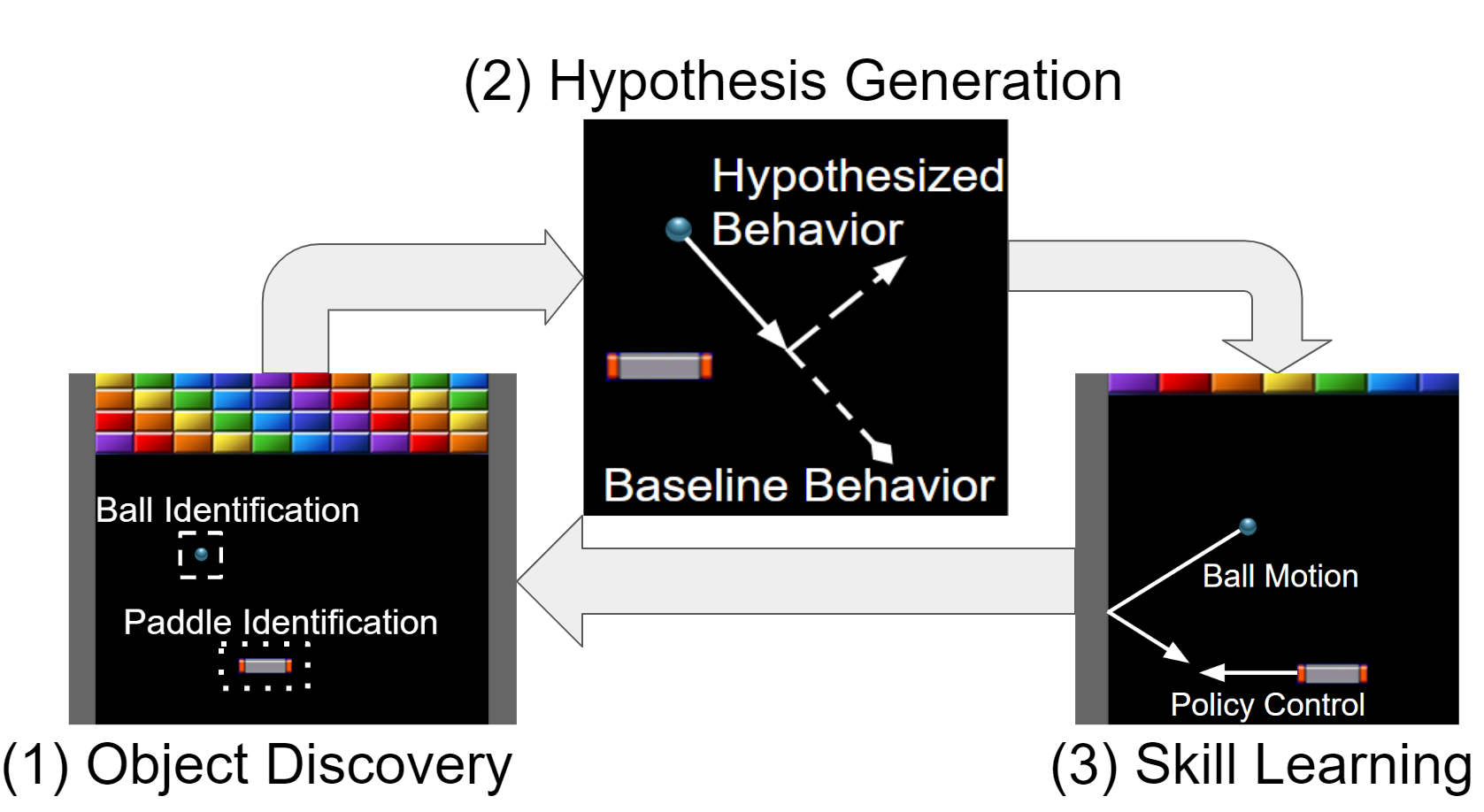}}
\caption{The HyPE loop illustrated on the benchmark RL task Breakout. (1) Beginning with primitive actions, controllable \textit{target} objects are visually discovered one at a time. (2) Hypotheses are generated that describe the types of changes that a \textit{source} object (or primitive action) may be able to cause in the target object. (3) Hypotheses are tested by attempting to learn \textit{options} that control the target object via the source object. Successfully learned options support learning in the next iteration of the HyPE loop.}
\label{HyPE}
\end{center}
\vskip -.9cm
\end{figure}

\begin{figure}
\vskip .3cm 
\begin{center}
\centerline{\includegraphics[width=0.95\columnwidth]{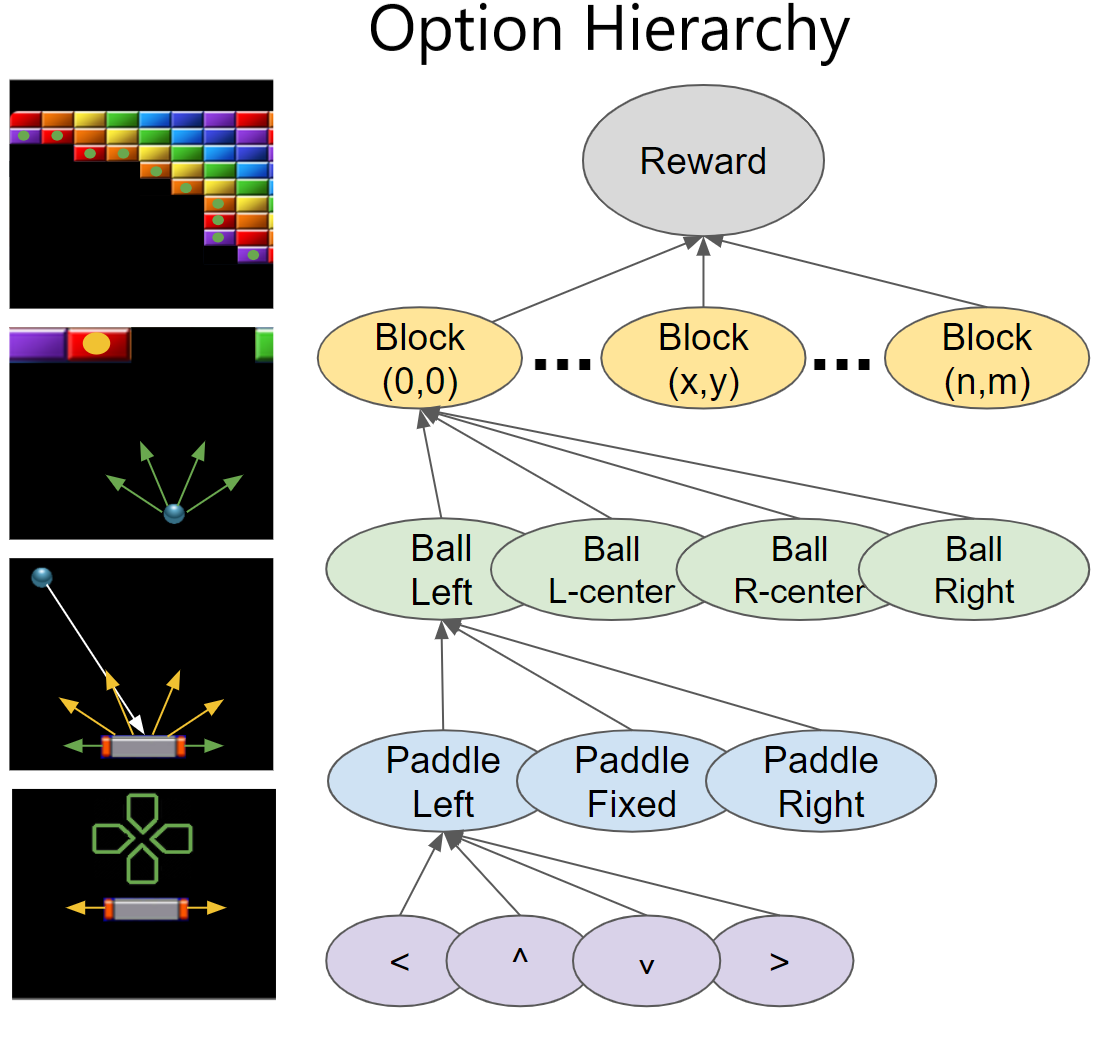}}
\caption{An example option hierarchy learnable by HyPE in Breakout. Each option's action space is comprised of options from the previous level, until reaching primitives actions at the bottom. 
An option that uses source object $o_i$ to control target object $o_j$ uses the object states $\phi_i(s), \phi_j(s)$ as input. 
Green in the images denotes the action space at each level, while yellow denotes the hypothesized target object state change(s), which define the termination sets for those options.}
\label{optionchain}
\vskip -1.0cm
\end{center}
\end{figure}

Thus, rather than working backwards from reward \cite{konidaris2009skill, florensa2017reverse}, we propose building forward towards state bottlenecks by learning skills that control sparse interactions between objects in physics-based domains. These skills can then be used to navigate bottlenecks, explore efficiently, and maximize return. In this work objects are physical entities that obey certain laws. Additionally, we add abstract objects such as primitive actions or reward that can affect, or be affected by, physical objects. 
In this work, if an action (either a primitive action or a higher-level option that controls an object) can cause a predictable change in another object, we refer to that action as being \textit{causal} of that change. The Hypothesis Proposal and Evaluation (HyPE) algorithm illustrated in Figure~\ref{HyPE} exploits causal relationships between objects using a three step loop to learn policies which navigate state space bottlenecks:

\begin{enumerate}
\topsep0pt
\itemsep0pt
    \item  {\bf Object discovery:} 
    In this step, we aim to discover factorized object representations from raw pixel inputs. Specifically, in each iteration of the HyPE loop, we attempt to discover one new object (the \textit{target} object) by learning a set of convolutional filters whose outputs meet certain physics-guided criteria, and whose motion is explained by a previously discovered object or primitive action (the \textit{source} object).
    
    \item {\bf Skill proposal via hypothesis generation:} 
    We then generate one or more hypotheses about the specific changes that the source object can cause in the target object by observing interactions between them (i.e. how the state of the source object appears to influence the state of the target object). However, these hypothesized interactions may be spurious rather than causal, so each hypothesis must be made testable. To do so, each hypothesis is instantiated as an option whose goal is to cause a particular change in the target object via the source object. The actions available to the option are, in turn, previously learned options that control the source object, beginning with primitive actions at the beginning of the hierarchy.

    \item {\bf Hierarchical skill learning via hypothesis evaluation:} 
    Finally, each interaction hypothesis is tested by using DRL to determine if the associated option is learnable. If an option policy can be successfully learned, the corresponding hypothesis is confirmed and the option hierarchy is extended permanently. The loop then continues from step 1. For example, in Breakout, options for controlling the ball can be learned by using options that control paddle displacement, which, in turn, use primitive actions. The next iteration of the HyPE loop could then use the new ball-control options to learn options which control the blocks. Figure \ref{optionchain} shows an illustration of such a hierarchy.

\end{enumerate}
We evaluate HyPE in two domains: First, a simulated robotic pushing domain in which standard DRL methods exhibit poor sample efficiency. Second, the classic arcade game of Breakout, where HyPE improves the sample efficiency of policy learning on raw pixels by an order of magnitude, as compared to several state-of-the-art DRL algorithms.
\section{Problem Formulation}
\subsection{MDP formulation}
A Markov Decision Process (MDP) is defined by $(S, A_\text{prim},T,R,\gamma)$. At each time step $t$, the agent observes a state $s \in S$ (in our case, $s$ is an image), with starting state distribution $S_0$, and takes primitive action $a \in A_{\text{prim}}$. The next state is determined by $T(s', a, s)$, the probability distribution over subsequent state $s'$ given current state $s$ and current action $a$.
The agent receives external reward as a function of the current state and action $R(s,a) \rightarrow \mathbb{R}$. The return $g_t$ is the discounted sum of rewards: $g_t = \sum_{i=t}^T\gamma^{i-t} R(s_i,a_i)$, where $\gamma$ is the discount factor. A policy $\pi(s|a)$ is defined the probability of an action $a$ given state $s$. Reinforcement learning searches for the policy that maximizes this total expected return. 

\subsection{Hierarchical Skills via Options}
Our skill hierarchy is based on the options framework \cite{options}. An option $\omega$ is defined by the tuple $(I_\omega, \pi_\omega, \beta_\omega)$, where $I_\omega$ is the initiation set, $\pi_\omega$ is a policy within the option, and $\beta_\omega : S\rightarrow [0,1]$ is the termination condition. We simplify the initiation set to say all options are available everywhere: $s \in I_\omega \forall s \in S, \forall \omega \in \Omega$. 

An option hierarchy \cite{konidaris} is a sequence of sets of options $\left[\Omega_1, \Omega_2, \hdots \Omega_n\right]$, where the action space for option $\{\omega_{k,i} \in \Omega_i\}$ are the options defined in $\Omega_{i-1}$, and the action space for $\Omega_1$ are the primitive actions. Thus, executing an option from $\Omega_n$ executes an option from $\Omega_{n-1}$, which itself executes an option from $\Omega_{n-2}$ and so on, until an option from $\Omega_{1}$ executes a primitive action.

The HyPE algorithm learns object specific option sets to learn a hierarchy of object control, treating primitive actions as the first object. Define an object state for object $o_i$ at time $t$ by the mapping $\phi_{o_i}(s^{(t)})$ from raw state to object factorized state. In this work, the object state is limited to an $(x,y)$ position, but this can be extended to any function of the state in future work. An object option set $\Omega_i$ contains $K$ options, $\omega_{k,i} \in \Omega_i$, where the terminal set of each option is
\begin{equation}
\beta_{k,i} = \{s^{(t)} , s^{(t-1)} | \phi_{o_i}(s^{(t)}) - \phi_{o_i}(s^{(t-1)}) \approx p_k\},
\end{equation}
and where $p_k$ is a single-step displacement corresponding to option $\omega_{k,i}$. The policy $\pi_{k,i}$ of this option uses options from $\Omega_{i-1}$ as the action space which are recursively defined as single step displacement of $\phi_{o_{i-1}}(s^{(t)})$.
For example, in breakout $\Omega_{\text{ball}}$ use $\Omega_{\text{paddle}}$ displacements as actions, and controls displacements in $\phi_{\text{ball}}(s^{(t)})$. We simplify $\phi_{o_i}(s^{(t)})$ to  $\phi_i^{(t)}$.
Figure~\ref{optionchain} describes this object option hierarchy for the Breakout domain.
\section{Methods}
In this section we introduce the HyPE loop, which at each iteration learns an object identification function $\phi_i(s)$ and adds an option set to an object option hierarchy $\mathcal H$, starting from $\Omega_\text{prim}$ and $\phi_\text{prim}$, the options and state space for primitive actions. $\phi_\text{prim}$ takes on the action on the current time step as state, and $\Omega_\text{prim}$ is the set of primitive actions. Each iteration performs three sub-steps. \textbf{1)} discovers a new target object by learning an object identification function $\phi_{\text{tar}}(s)$. This function tracks the object in the scene, learning $\phi_\text{tar}$ using correlations with a source object $o_{i}$. \textbf{2)} proposes hypotheses about how the object can be controlled using $\Omega_i$, the options controlling the source object. \textbf{3)} uses deep reinforcement learning to learn these options which produce the proposed control as the termination set $\beta_\text{tar}$ of that new option. If the policy achieves non-trivial reward, HyPE adds $\Omega_\text{tar}$ to $\mathcal H$ and keeps track of $\phi_\text{tar}$. The HyPE loop then iterates again using $\phi_\text{prim}, \hdots, \phi_\text{tar}$  to discover additional new objects, terminating when it has achieved high task reward (The object option hierarchy $\mathcal H$ is task specific). See Algorithm~\ref{hypealgo}.
 
\begin{algorithm}[H]
  \caption{HyPE Loop}
  \label{hypealgo}
\begin{algorithmic}
  \STATE {\bfseries Input:} MDP Environment $E$
  \STATE {\bfseries Initialize} Object option hierarchy $\mathcal H$ with a single option set $\Omega_{\text{prim}}$, $\phi_\text{prim}$
  \STATE {\bfseries Data} $\mathcal D = \mathcal D \bigcup \text{random policy data}$
  \REPEAT
      \STATE {\bfseries Object Discovery}: learn $\phi_{\text{tar}}(s)$, where the $\{\phi_{\text{tar}}^{(t)}\}_{t\in D}$ interacts with a source object $\phi_{i}$, by optimizing Equation ~\ref{visionloss} (Section III.A) with black box optimization
      \STATE {\bfseries Hypothesis Proposal}: Propose $\{p_{k, \text{tar}}\}_{k=1}^K$ which define the termination conditions $\beta_{k, \text{tar}}$ for options controlling $o_{\text{tar}}$. (Section III.B)
      \STATE {\bfseries Hypothesis Evaluation}: Learn the policies $\pi_{k,\text{tar}}$ for the new set of options using Deep reinforcement learning. (Section III.C)
      \STATE {\bfseries Update} Add $\phi_{\text{tar}}, \Omega_{\text{tar}}$ to $\mathcal H$ if the policy in Hypothesis Evaluation is learnable
  \UNTIL{A HyPE option achieves high extrinsic reward}
\end{algorithmic}
\end{algorithm}

\subsection{Object Discovery}
The object discovery step learns an object identification function---the mapping from images to object state $\phi_{\text{tar}}(s)$---for target object $o_{\text{tar}}$.
In order to learn this function, we use the following insight: we can discover new objects by tracking interactions with a source object $o_i$ learned in a previous iteration of the HyPE loop. In practice, we represent the object identification function $\phi_{\text{tar}}(s)$ as a convolutional neural network (CNN) which outputs a heatmap over the input image $s$, and returns the $(x,y)$ pixel coordinate with highest intensity response.

For the object discovery step we optimize a loss function over a sequence of object states of source object $\phi_i$ and the target object (being learned) $\phi_\text{tar}$ where $\vec{\phi}_k = (\phi_k^{(t)})_{t=0}^T$ 
\begin{equation}
\begin{split}
L(\vec{\phi}_i; \vec{\phi}_{\text{tar}})
    = &-F_1(\vec{\phi}_i; \vec{\phi}_{\text{tar}}) + \lambda_1 \sum_{t=0}^{T}\left\|\phi_{\text{tar}}^{(t+1)} - \phi_{\text{tar}}^{(t)}\right\|
\end{split}
\label{visionloss}
\end{equation}
The $F_1$ criteria measures the relevance of \textit{correlated interactions} between $\phi_\text{tar}$ and source object state $\phi_i$. A correlated interaction is when there is a \textit{changepoint}, when the motion of the target object changes significantly, during an \textit{eligible} time, which is when two objects are likely to be interacting. We will define these formally. The second term penalizes the l2 distance between sequential outputs by $\phi_{\text{tar}}$. $L(\vec{\phi}_i; \vec{\phi}_{\text{tar}})$ is optimized using a black box optimization algorithm, CMA-ES, over the weights of the CNN.



To define \textit{eligible}, we use spatial proximity as a physical heuristic for the ability of objects to interact. Since our state is defined by $(x,y)$ coordinates of the objects, two objects are eligible when 
\begin{equation}
\|\phi_{\text{tar}} - \phi_i\| \leq \epsilon_{\text{prox}}
\label{proximity}
\end{equation}
The $\epsilon_{\text{prox}}$ hyperparameter specifies a pixel distance threshold. This depends on the geometry of the objects, so though ideally we would define the distance in terms of edges or nearest point analysis, for the purpose of simplicity, in this paper we use point distance. We define $t$ as an \textit{eligible time} if $\phi_{\text{tar}}^{(t)}, \phi_{i}^{(t)}$ satisfy Equation~\ref{proximity}.

We detect \textit{changepoints} using Changepoint Detection using Approximate Model Parameters (CHAMP)~\cite{niekum}. Changepoints $\{c_1, \hdots, c_M\}$ 
are timesteps in a trajectory. Each pair of sequential time steps $c_m, c_{m+1}$ defines a segment $[\phi_\text{tar}^{(t)}]_{t \in [c_m \hdots c_{m+1}]}$, where a model $q_{m}(\phi_\text{tar}^{(t)}) \in Q$ approximates the state transition $q_{m}(\phi_\text{tar}^{(t)}) \approx \phi_\text{tar}^{(t+1)}$,
We use the affine model class for $Q$, such that $q(\phi_\text{tar}^{(t)}) = D\phi_\text{tar}^{(t)} + d \approx \phi_\text{tar}^{(t+1)}$. D is an $2\times 2$ matrix and $d$ is a length 2 vector, learned by linear regression. The number of changepoints $M$ is discovered from data. With abstract objects like primitive actions, where proximity does not make sense as a criteria for eligible, we use changepoints in the object $o_i$ as eligible times. 

When an eligible time co-occurs with a changepoint, we call this a correlated interaction. The F1 score measures the the harmonic mean of precision and recall between eligible times and changepoints. The intuition is that for the source object state to be correlated with the target object state, there should be more changepoints than usual when eligible---otherwise the changepoints can be seen to be independent of the source. Using the F1 score to quantify correlated interactions balances eligible times with changepoints. Without balancing, the CNN might learn to track the source object (always eligible) or constantly exhibit difficult to model displacements (many changepoints).

Given a trajectory $[\phi_i^{(t)}, \phi_\text{tar}^{(t)}]_{t=0}^T$, we can define length $T$ binary vectors for eligible times $\textbf{e}$ and changepoints $\textbf{c}$, which are $1$ at an eligible time/changepoint respectively, and $0$ elsewhere. The F1 score for correlated interactions is
\begin{equation}
F_1(\vec{\phi}_i; \vec{\phi}_{\text{tar}}) = \frac{1}{\frac{\|\textbf{c}\|_2^2}{\textbf{c}^\top \textbf{e}} + \frac{\|\textbf{e}\|^2_2}{\textbf{c}^\top \textbf{e}}}
\end{equation}
This is just one possible instantiation of eligibility and changepoints---future work could use heuristics or learned metrics.

The score achieved by the F1 component of $L(\vec{\phi}_i; \vec{\phi}_{\text{tar}})$ provides a measure used to verify that object discovery has learned an object that is likely to be controllable through $o_i$. If the final F1 score after learning is low, this means that the learned $\phi_{\text{tar}}(s)$ does not map to a feature that has many correlated interactions $o_i$, and might be noise. This is a good criteria for deciding when object discovery should move to a different source object $o_j$. In Breakout, for example, the abstract object for primitive actions $\phi_{\text{prim}}$ has high F1 score when locating the paddle, because changing actions is highly correlated with changes in paddle motion. After the paddle is learned, however, the F1 score for primitive actions in the scene is low. This stopping criteria is:
\begin{equation}
F_1(\vec{\phi}_i; \vec{\phi}_{\text{tar}}) \leq \epsilon_{F1}
\label{visiontesteq}
\end{equation}

Though the vision system is sufficient to achieve the results we describe in Section IV, we acknowledge some shortcomings: first, the loss is defined pairwise between only two objects, meaning that objects which have multiple simultaneous interactions are difficult to identify. Second, it is necessary to remove already learned objects from $s$, or they might be learned repeatedly. We do this by subtracting the mean of a fixed region around an already learned object from $s$, which masks out the learned object.
Finally, the vision loss uses a single image as input into a CNN of limited size (due to the computational cost of running CMA-ES, even for moderately sized neural networks). This will be addressed in future work by taking in multiple frames as input and designing a differentiable definition of eligibility and changepoints. 


\subsection{Hypothesis Proposal}
Object discovery determines that $\phi_\text{tar}$ has correlated interactions with $o_i$, but does not specify how to learn options to manipulate $\phi_\text{tar}$. Hypothesis generation constructs the set of possible motions, the hypotheses, that $o_i$ can effect on $\phi_{\text{tar}}$. Each proposed motion $p_{k,\text{tar}}$, is a hypothesis about the way that $o_i$ can causally manipulate $\phi_{\text{tar}}$. Learning these options in hypothesis evaluation would verify that $o_i$ causes this desired motion in  $\phi_{\text{tar}}$.


Hypothesis proposal uses distinct motions $p_{k,\text{tar}}$ to define the termination set for option $\omega_{k,\text{tar}} \in \Omega_{\text{tar}}$. Motion is represented as a single timestep change in $\phi_\text{tar}$: $\phi_{\text{tar}}^{(t+1)} - \phi_{\text{tar}}^{(t)}$. For states $s^{(t)}$ followed by $s^{(t+1)}$, the termination set $\beta_{k,\text{tar}}$ corresponding to hypothesized motion $p_{k,\text{tar}}$ has the form
\begin{equation}
\beta_{k, \text{tar}} = \{s^{(t)},s^{(t+1)}|\|\phi_{\text{tar}}(s^{(t+1)}) - \phi_{\text{tar}}(s^{(t)}) - p_{k,\text{tar}}\| \leq \epsilon_{k, \beta}\}
\label{terminationset}
\end{equation}



To define $K$ motions $\{p_1 \hdots p_K\}$, we want to specify as few motions as possible, while capturing all single timestep motions over $\phi_\text{tar}$ caused by $o_i$. Limiting the number of $p_{k,\text{tar}}$ reduces the cost of policy learning in hypothesis evaluation. Thus, even though the set of $p_k$ could be all observed single timestep displacement after a correlated interaction (the set of correlated actions being $CI = \{t | \textbf{e}(t) \cdot \textbf{c}(t)=1\}$): $\hat P = \{\phi_\text{tar}^{(t+1)} - \phi_\text{tar}^{(t)}\}_{t\in CI}$, this set would include many spurious changepoints due to multi-object interactions and vision failures.
Instead, to reduce noise, take $\hat P$ 
as the training set for a DP-GMM (Dirichlet process Gaussian mixture model)~\cite{escobar1994estimating} an unsupervised clustering model, and take clusters with the number of data points assigned to them above a fixed minimum $n_{\text{DPGMM}}$. These cluster means are used as parameters $p_{k,\text{tar}}$. The $\epsilon_{k, \beta}$ are computed using the cluster variance of the Gaussian model corresponding to the respective cluster. This unsupervised method for discretizing and denoising object changes is one choice of method for defining the set of hypothesized control.

For the hypotheses: $o_i$ causes $\{p_{1, \text{tar}} \hdots p_{K, \text{tar}}\}$, we not only want to propose a set of possible motions, but also ensure that this set of motions is caused by $o_i$. To do this, we specify the input space and (output) action space of the policy which will be learned to fulfill the termination condition. 
Thus, the input state of the target options is the state of the source object $\phi_{i}$ and the target object $\phi_{\text{tar}}$ only. For example, a paddle-ball policy ignores block state. The action space is $\Omega_i$, or options manipulating $o_i$ by $o_{i-1}$. This ensures that the option manipulates the target object $o_{\text{tar}}$ via the source object $o_i$, and is blind to objects other than the two it is learning the interaction between.

In summary, hypothesis proposal generates a set of possible options $\Omega_\text{tar} := \{\omega_{1,\text{tar}}, \hdots, \omega_{K,\text{tar}}\}$ distinguished by a particular displacement motion that they effect on $\phi_\text{tar}$, using $\Omega_i$ as actions and ignoring all state except $[\phi_i, \phi_\text{tar}]$.



\subsection{Hypothesis testing to determine causal links}
Hypothesis testing differentiates correlation between two objects, as observed in the previous two steps, and causal relations, by attempting to learn the policies $\{\pi_{1,\text{tar}}, \hdots, \pi_{K,\text{tar}}\}$ corresponding to the proposed option set $\Omega_\text{tar}$ using DRL.

Using DRL requires a reward function. We convert the termination condition of the option into an intrinsic reward for training the policy by: 
\begin{equation}
R(s,s',p_{k,\text{tar}}) = E(\phi_\text{tar}, \phi_i)C(\phi_\text{tar})B_{k}(\phi_\text{tar},\phi_\text{tar}', p_{k,\text{tar}})
\label{RewardFunction}
\end{equation}
E, C, B are a 0-1 indicator functions of eligibility, changepoints and termination set $\beta_{k,\text{tar}}$ respectively (Equation~\ref{proximity}, Section III.A, Equation~\ref{terminationset}). This gives nonzero reward of $1$ only for correlated interactions which produce the desired single timestep displacement $p_{k,\text{tar}}$. 

We use DRL with this reward function and test a variety of different DRL algorithms, including actor critic (PPO) ~\cite{schulman}, Actor (policy iteration), Critic (Q-learning) ~\cite{mnih} and black box (CMA-ES) ~\cite{hansen}. 

In order to exploit the object factorized structure of $[\phi_i, \phi_\text{tar}]$, we utilize a neural net which computes:
\begin{equation}
\text{softmax}\left(\textbf{W}_{\text{ff}}\frac{1}{L}\sum_{l=0}^L \sigma(\textbf{W}_{\text{emb}}\psi_l(\phi_i, \phi_\text{tar}))\right)
\end{equation}
Where $\psi_l(\cdot, \cdot)$ are basic input features computed from $\phi(\cdot)$, such as relative position or velocity, while still including $\phi_i$ and $\phi_{\text{tar}}$. This network then expands each input feature into a length $N$ embedding, where $W_{\text{emb}}$ is a $N\times 2$ matrix of weights (all input features have dimension 2), and $\sigma$ is the rectified linear unit. It takes the mean of all $L$ embeddings vectors and feeds these forward to action logits, using a softmax operation to convert these to probabilities. 

To train the $K$ options simultaniously, corresponding to each $p_{k,\text{tar}}$, we randomly switch between executing $\pi_{k,\text{tar}}$ for a fixed duration, and perform off-policy updates when amenable. When using an on-policy DRL algorithm, then we update on-policy. 

Hypothesis evaluation assesses the causal relation between $o_i, o_{\text{tar}}$ by comparing the expected return of the learned policy $\mathbb E[g_t|\pi_{k,\text{tar}}]$ (computed using the intrinsic reward in Equation~\ref{RewardFunction}), with the expected return of a random policy $\mathbb E[g_t|\pi_\text{random}]$. 
\begin{equation}
\mathbb E[g_t|\pi_{k,\text{tar}}] - \mathbb E[g_t|\pi_\text{random}] \geq \epsilon_\text{causal}
\label{hypothesistest}
\end{equation}
A policy which satisfies the above criteria produces the hypothesized state change $p_k$ more often than a random policy. This is a usable option for manipulation of $\phi_{\text{tar}}$, so we add $\Omega_\text{tar}$ to the $\mathcal H$. 

Since $\pi_{k, \text{tar}}$ uses option set $\Omega_{i}$ to manipulate $\phi_{\text{tar}}$, and $\pi_{k, \text{tar}}$ has a limited input space only including $[\phi_i, \phi_{\text{tar}}]$, this policy can be seen as an intervention where $o_i$ causally controls $o_\text{tar}$. In addition, learning at least one option $\omega_{k, \text{tar}}$ demonstrates some control over $o_\text{tar}$ by the agent, since the base node of the object option chain is primitive actions.

\subsection{Overall HyPE Loop}

The HyPE algorithm applied to a new domain, as described in Algorithm~\ref{hypealgo}, repeatedly loops between object discovery, hypothesis proposal and hypothesis evaluation. In order to begin object discovery, historical data $\mathcal D$ is initialized by collecting data from a policy which takes random actions. This sample also forms the baseline comparison for Equation~\ref{hypothesistest}. 

The only option set in $\mathcal H$ on the first iteration of the HyPE loop corresponds to primitive actions. $\phi_{\text{prim}}$ collects the actions taken, and $\omega_{k,\text{prim}}$ corresponds to the the $k^{\text{th}}$ primitive action. Option discovery optimizes Equation~\ref{visionloss} with $o_i=o_{\text{prim}}$ using trajectories from $\mathcal D$ as training data. The object identification function $\phi_1(s)$ returns a $\phi_1$ with correlated interactions with $A_{\text{prim}}$. Then, hypothesis proposal and evaluation will learn the option set $\Omega_1$ to manipulate $\phi_1$, and add $\Omega_1$ to $\mathcal H$.

The subsequent iterations of the loop will add new option sets to the chain from the bottom up. The loop starts object discovery with $o_\text{prim}$ as the source object. If optimizing Equation~\ref{visionloss} fails to learn a $\phi_\text{tar}$ that satisfies Equation~\ref{visiontesteq}, or the hypothesis evaluation fails (Equation ~\ref{hypothesistest}), the object discovery restarts with $\phi_1$. This traversal heuristic assumes that objects more directly manipulated will be easier to learn about. In the case of multiple chains (i.e. multiple objects are directly manipulated by some $o_i$), HyPE finds the shortest chain to manipulate reward by discovering objects in a breadth-first style search.

\section{Results}
We demonstrate the capabilities of the HyPE algorithm in two domains. First, we show HyPE learns policies which achieve good performance in a perfect perception robotic pushing domain, where classical reinforcement learning methods perform poorly. In the robotic domain, common DRL baselines struggle to learn fairly intuitive policies because of the state space bottlenecks in pseudo-physical domains. We also show HyPE learns high scoring policies from pixels in a sample efficient manner in the classic game (and deep DRL benchmark task) Breakout, where standard deep DRL policies take many more timesteps.
\subsection{Robotic Pushing Domain}
In Our 2-D robotic pushing domain, a gripper, controlled in cardinal directions, manipulates a block by pushing it into a target location. The agent receives non-zero reward of 1 only if the block contacts a target area. Episodes end when the block contacts the target area, or after 300 timesteps. All three objects have randomly initialized positions. 
This domain is challenging for standard RL because the reward is extremely sparse: a random policy takes on average $10,000$ time steps to stumble upon non-zero reward. 

In the pushing domain, we seek to demonstrate that HyPE provides clear benefits beyond factorized state. Thus, the robot pushing domain has perfect perception. Incorporating perception with the full HyPE loop is the focus of the Breakout experiments. The state consists of three pairs of $(x,y)$ pixel coordinates corresponding to the gripper, block, and target. HyPE iterations in the robotic pusher domain start with hypothesis proposal to learn relations between the paired positions. Hypothesis evaluation learns options that perform the desired behavior, utilizing Proximal Policy Optimization (PPO)~\cite{schulman} with the 0-1 reward and $W_{\text{emb}}$ with $N=4096$ (as defined in Section III.C). 

\begin{table}[t]
\vskip 0.2cm
\caption{Table of training times for each loop of HyPE when learning the robotic pushing domain. ``Random'' stands for the initial random states to start the loop, while ``gripper'', ``block'' and ``reward'' denote HyPE iterations learning to control each of these objects.}
\label{tableRPD}
\begin{center}
\begin{small}
\begin{sc}
\begin{tabular}{lcccr}
\toprule
Random & Gripper & Block & Reward \\
\midrule
$2,000$ & $2,000$ & $1,500,000$ & $1,000,000$ \\
\bottomrule
\end{tabular}
\end{sc}
\end{small}
\end{center}
\vskip -0.2in
\end{table}
\begin{figure}[t]
\begin{center}
\centerline{\includegraphics[width=0.87\columnwidth]{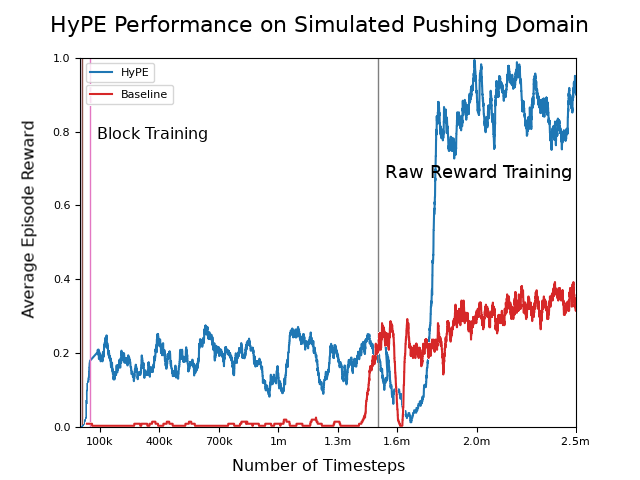}}
\caption{Graph of training in the robotic pushing domain. Baseline methods A2C, PPO and Rainbow failed to achieve better reward than random actions, even with reward shaping. A reward-hacked baseline (red) is able to achieve better than trivial rewards, but worse performance than HyPE. The vertical lines dictate different iterations of the HyPE loop.}
\label{pusherplot}
\end{center}
\vskip -1.0cm
\end{figure}
The HyPE loop begins with only $\phi_\text{prim}$ and $\Omega_\text{prim} \in \mathcal H$, and initializes the dataset with random actions until it picks up a correlation between primative actions and another state variable, which takes $\sim1000$ random steps. HyPE proposes 
five motions over the gripper ($p_{k, \text{gripper}}$),  left, right, up, down, and stationery, with $\pm \sim2.4$ pixels as cluster means of the learned DP-GMM. Hypothesis evaluation learns a policies corresponding to these options in $\sim5000$ time steps. 

The data from this step allows hypothesis proposal about block control. However, because block changepoints occur infrequently, the initial options only capture limited control of the block, without rewarding all directions. After $\sim10,000$ time steps, however, the options are re-specified with left, right, up, down motions, four $p_{\text{block}, k}$. HyPE learns these policies in an average of $1.5M$ time steps of training, leading to policies which push the block in the desired direction.

The third iteration of the HyPE loop reveals that the relationship between the block and target is correlated with extrinsic reward (and end of the episode). HyPE proposes options to control the extrinsic reward by controlling the block. Since this is a multi-object interaction, where a changepoint in extrinsic reward based on the relative positions of block and target. We augmented HyPE hypothesis proposal by allowing proximity between $\phi_\text{block}, \phi_\text{target}$ (or any other pair of objects) to be correlated with changepoints in $\phi_\text{reward}$. The policy for controlling extrinsic reward learns in an average of $500k$ time steps. Figure~\ref{pusherplot} shows the full performance of HyPE in the pushing domain. Learning the reward option set requires fewer time steps than the block option set, because HyPE uses control of the block as the action space for this option, and only needs to plan using the block-target relative information. 

By comparison, A2C \cite{mnih}, Proximal Policy Optimization (PPO) and Rainbow~\cite{hessel} trained on the same domain, using the baseline reward, return policies that do not perform better than random even after $20M$ time steps. Even when given a shaped reward, which is equal to $-\lambda \|\phi_{\text{Block}} - \phi_{\text{Target}}\|_1$, a scaled negative l1 norm shaped reward between the block and the target, these standard RL algorithms fail to learn meaningful policies. We constructed one baseline which succeeded in pushing the block to the goal approximately $60\%$ of the time after $15M$ time steps (as compared to the $90-95\%$ success rate from HyPE). The specialized HyPE-like reward gave $+1$ reward for moving the block, and a $+1000$ reward for end of episode. This demonstrates not only that HyPE learns a very reasonable set of sub-tasks, but also that the action hierarchy of HyPE, the main difference between this baseline and HyPE, is invaluable in solving some tasks. This results overall demonstrate how the object option chain from HyPE can be used to solve problems which would by standard RL be infeasible. Table~\ref{tableRPD} shows the timesteps needed for learning to control the different objects.

\subsection{Breakout Domain}
\begin{figure*}
\vskip .15cm
\centering
\begin{subfigure}{.4\textwidth}
  \centering
  \includegraphics[width=1\columnwidth]{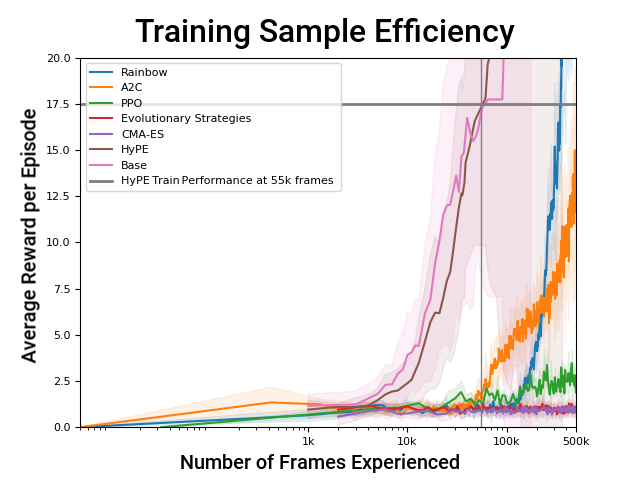}
\end{subfigure}%
\begin{subfigure}{.4\textwidth}
  \centering
  \includegraphics[width=1\columnwidth]{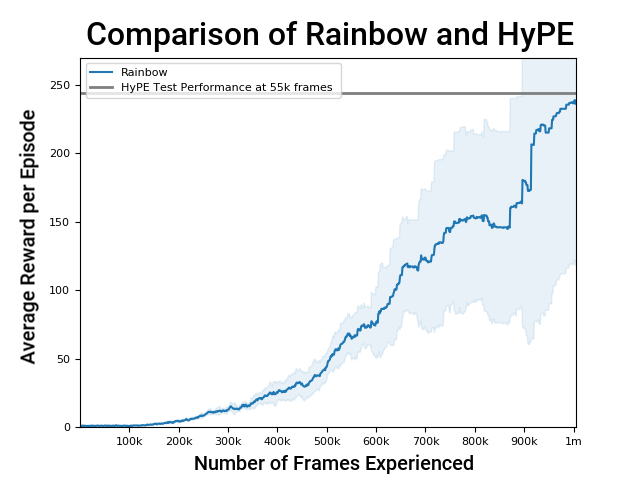}
\end{subfigure}
\caption{Performance comparison (average of 10 trials) between PPO, A2C (orange), Rainbow (blue) and HyPE (maroon), evolutionary strategies \cite{salimans}, CMA-ES and a perfect perception, paddle-ball state only baseline (Base). The y-axis is average episode return, and the x-axis is number of frames experienced, on a log scale. HyPE outperforms Rainbow~\cite{hessel}, the best baseline not using perfect perception, by 7x in training and 18x in test. HyPE testing performance at 55k frames matches Rainbow performance after 1m time steps (see Table~\ref{table}). Since the Baseline is a perfect reward, perfect perception equivalent of HyPE, we expect it to lower bound HyPE performance.}
\label{plot}
\end{figure*}
\begin{table}[t]
\vskip .2cm
\caption{Table of training time to find evaluation policy with 244 blocks hit in Breakout, the average test score of HyPE after 55,500 frames of training (standard error 27, 20 trials). ``Base'' is a CMA-ES algorithm run on the relative positions of the paddle and the ball, ball velocity, and ball and paddle positions, from the true underlying game state.}
\label{table}
\begin{center}
\begin{small}
\begin{sc}
\begin{tabular}{lcccr}
\toprule
Base & HyPE & Rainbow & A2C \& PPO \\
\midrule
52,000 &55,500& $\sim1,000,000$ & $>1,500,000$\\
\bottomrule
\end{tabular}
\end{sc}
\end{small}
\end{center}
\vskip -0.4cm
\end{table}

We add object discovery to the HyPE loop in Breakout (Figure~\ref{HyPE}). As before, starting from only $\phi_\text{prim}$ and $\Omega_{\text{prim}}$, HyPE takes random actions until it discovers an object correlated with primitive actions. After $\sim1000$ frames of random data, HyPE learn a object detection CNN to locate the paddle by optimizing the loss defined in Equation~\ref{visionloss}. With sufficient F1 score to pass Equation~\ref{visiontesteq}, the loop proposes and learns to control the paddle $\pm 2$ or $0$ pixels in 2.5k timesteps. The HyPE loop adds object identification function $\phi_\text{paddle}$ and option set $\Omega_\text{paddle}$.

Using the cumulative data, the HyPE loop then optimizes the F1 score starting with $o_\text{prim}$ as the source object. With the paddle removed from the image, the F1 score does not pass Equation~\ref{visiontesteq}. However, using $o_\text{paddle}$, object discovery learns a CNN which tracks the ball.
Due to the rarity of ball bounces, the proposed option only bounces the ball off the paddle---it groups all angles together, leaving a single ball control option. Learning this option is sufficient to control extrinsic reward, so HyPE terminates.

In Figure~\ref{plot}, We show that HyPE has an order of magnitude improvement in sample efficiency when compared to standard RL methods on Breakout. HyPE, at train time, achieves average train reward per episode of $17.5$ in 55k frames, while Rainbow~\cite{hessel} takes 400k timesteps, and Proximal Policy Optimization~\cite{schulman} and A2C \cite{mnih} take roughly 1.4M timesteps to achieve the same performance. However because CMA-ES, which used to learn the HyPE policies in Breakout, typically has substantially higher test than train performance, comparing training performance understates the performance of the learned policy. The evaluation policy learned by HyPE after 55k frames achieves $244$ average reward per episode performance. Rainbow, the best performing baseline, takes $\sim1M$ timesteps to achieve the same performance. 

Note that though the HyPE loop learns intuitive objects (the paddle, ball), this is not encoded explicitly anywhere in the algorithm but emerges from physical priors and controllability. HyPE has the same information as the standard RL algorithms, but uses object priors to achieve high sample efficiency.
\section{Related Work}
This work combines ideas from causality and relational learning, model-based reinforcement learning, and hierarchical reinforcement learning. It uses these ideas to construct a hierarchical reinforcement learning problem with intrinsic rewards. 

\textbf{Causal graphs:} The hypothesis proposal and evaluation ideas draw from causality literature \cite{pearl}. These components of HyPE relate to where causal graphs \cite{Shanmugam} and graph-dependent policies \cite{Buesing} learned from interactions with the environment. HyPE also learns causal object relationships \cite{battaglia, garnelo, toyer, Kansky} which has similarities to object-oriented and relational reinforcement learning \cite{diuk, zambaldi, silva, watters}. Though HyPE uses a similar object-oriented relational structure, it learns object-object interactions one at a time for highly efficient hierarchical reinforcement learning and visual factorization. 

\textbf{Model-based RL:} Model-based reinforcement learning using a learned model has shown substantial improvements in reinforcement learning sample efficiency in Atari games \cite{kaiser}. These methods often learn to predict future raw state \cite{silver2017predictron}, or latent space \cite{watter2015embed} in tandem to learning a useful policy \cite{schmidhuber2015learning}. They then incorporate planning \cite{piergiovanni2018learning, lowrey2018plan, nagabandi2018neural} with constructed environment models \cite{racaniere2017imagination}. HyPE makes loose use of modeling to generate different options, but once it learns control policies, it applies model-free reinforcement learning and should be improved by incorporating Model-based RL.

\textbf{Hierarchical RL and Exploration:} Learning hierarchies of control with options has been studied in detail \cite{sutton, bacon}, and can be used to define a system for learning skills and state spaces \cite{konidaris, kulkarni2016hierarchical, daniel2012hierarchical, barreto2017successor, vezhnevets2017feudal, arulkumaran2016classifying}. Exploration work has used novelty \cite{ostrovski, Tang, burda}, frontier states \cite{savinov, ecoffet}, model prediction error \cite{burda2018large, pathak2017curiosity}, sub-goals from hindsight \cite{andrychowicz}, bottleneck regions \cite{bacon} or contingency \cite{bellemare, choi} as other exploration objectives. While the HyPE loop is inspired by these works, it incorporates object option hierarchies and physical priors.
\section{Conclusion}
We introduced the HyPE algorithm, which explores physically inspired state space bottlenecks to efficiently learn to hierarchically explore and control its environment. By taking advantage of some physically inspired priors like proximity and changepoints, this system learns high performing policies in RL settings. Though this system is less application agnostic than classic general-purpose RL algorithms, it achieves sample efficiency that is an order of magnitude better than standard RL methods on multiple domains. Future work can address the practical issues required to extend HyPE to physical real-world domains. Furthermore, the object option chain structure generated by HyPE may have implications for both explainable AI and transfer learning.

\bibliography{ICRA_2020}
\bibliographystyle{ieeetr}

\end{document}